\documentclass[runningheads]{llncs}
\usepackage{graphicx}

\usepackage{tikz}
\usepackage{comment}
\usepackage{amsmath,amssymb} 
\usepackage{color}

\usepackage{graphicx}
\usepackage{multirow}
\usepackage{wrapfig}
\usepackage{mathtools}
\DeclarePairedDelimiter\ceil{\lceil}{\rceil}

\usepackage[accsupp]{axessibility}  

\begin{document}
\pagestyle{headings}
\mainmatter
\def\ECCVSubNumber{7384}  

\title{IGFormer: Interaction Graph Transformer for Skeleton-based Human Interaction Recognition} 

\titlerunning{IGFomer}
%
\author{Yunsheng Pang\inst{1} \and
Qiuhong Ke\inst{1,2} \thanks{Corresponding author} \and
Hossein	Rahmani\inst{3} \and 
James Bailey\inst{1} \and
Jun Liu\inst{4}
}
\authorrunning{Y. Pang et al.}
%
\institute{The University of Melbourne, Australia \\
\email{yunshengp@student.unimelb.edu.au}, \email{baileyj@unimelb.edu.au} \and
Monash University, Australia \\
\email{Qiuhong.Ke@monash.edu} \and
Lancaster University, United Kingdom\\
\email{h.rahmani@lancaster.ac.uk} \and
Singapore University of Technology and Design, Singapore\\
\email{jun\_liu@sutd.edu.sg}}
\maketitle

\begin{abstract}
Human interaction recognition is very important in many applications. One crucial cue in recognizing an interaction is the interactive body parts.
In this work, we propose a novel Interaction Graph Transformer (IGFormer) network for skeleton-based 
interaction recognition via modeling the 
interactive body parts 
as graphs. More specifically, the proposed IGFormer constructs interaction graphs according to the semantic and distance correlations between the interactive body parts, and 
enhances the representation of each person by aggregating the information of the interactive body parts based on 
the learned  graphs. Furthermore, %
we propose a Semantic Partition Module to transform each human skeleton sequence into a  
Body-Part-Time sequence to better capture the spatial and temporal information of the skeleton sequence for learning the graphs. 
Extensive experiments on three benchmark datasets demonstrate that 
our model 
outperforms
the state-of-the-art with a significant margin.

\end{abstract}

\section{Introduction}


Human interaction recognition plays a significant role in a wide range of applications ~\cite{AggarwalRyoo11csur,dcf86615e06947bcb3f2c3eccb615857,zhang2012spatio,vahdat2011discriminative}. For example, it can be used in visual surveillance to detect dangerous events such as ``kicking'' and ``punching''. It can also be used for robot controlling for human-robot interaction. This paper addresses  human interaction recognition from skeleton sequences ~\cite{Shahroudy_2016_CVPR,liu2019ntu}. Compared with RGB videos, skeleton sequences provide only 3D coordinates of human joints, which are more  robust to unconventional and variable conditions, such as unusual viewpoints and cluttered backgrounds.


\begin{figure}[t] 
    \centering 
    \includegraphics[width=0.7\textwidth]{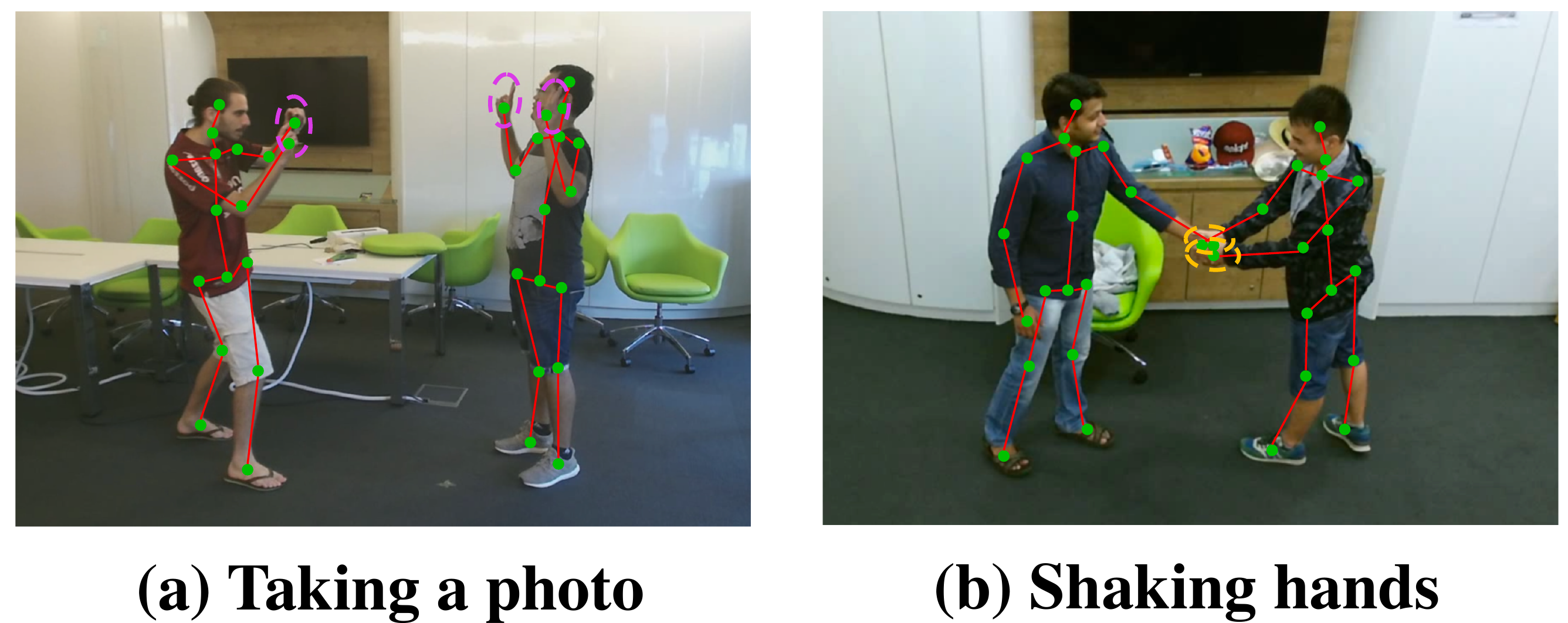} 
    \caption{(a) In the interaction of ``Taking a photo'', 
    there is strong semantic correlation between the hands holding the camera of one person and the hands with ``yeah" of the other person. (b) In the interaction of ``Shaking hands'', the interactive body parts  demonstrate both semantic correlation and distance evolution, i.e., the hands of two interactive persons correspond to each other and are gradually close to each other when they are shaking hands.
    } 
    \label{moti}  
    \vspace{-0.5cm}
\end{figure}

Compared with single-person action recognition, one additional crucial cue in recognizing a human interaction is the interactive body parts of the interactive persons. For example, the interactive hands of two persons are 
critical in understanding a ``shaking hands'' interaction. Generally, the interactive body parts in interactions demonstrate  semantic correlations and correspondence. For example, in the interaction of ``Taking a photo'' shown in Fig.~\ref{moti} (a), the hands holding the camera of one person and the hands with ``yeah'' of the other person demonstrate a strong correlation. Similarly,
in ``Shaking hands'' shown in  Fig.~\ref{moti} (b), the interactive hands of the two persons correspond to each other.
  In these cases,
exploring the semantic correlation between the interactive body parts 
is  crucial for interaction understanding.
In addition,
for some interactions, the interactive body parts demonstrate distance evolution.
For example, 
the  hands of 
the two persons gradually approach each other when the two persons are ``shaking hands''.
Measuring the  distance between body parts of the interactive persons can provide additional useful information to the semantic correlation for better interaction recognition.

Inspired by the above observation and the successful application of Transformer in many fields \cite{devlin-etal-2019-bert,dosovitskiy2020image,zhu2021deformable,SETR}, 
we propose a novel Transformer-based model named Interaction Graph Transformer (IGFormer) 
for interaction recognition from skeleton sequences. In particular, the proposed IGFormer consists of a  Graph Interaction Multi-head Self-Attention (GI-MSA) module, which aims at modeling the  relationship of 
interactive persons from both 
semantic and distance levels to recognize actions. 
More specifically, the GI-MSA module learns a semantic-based graph and a distance-based interaction graph to represent the mutual relationship between body parts of the interactive persons. 
The semantic-based graph is learned by the attention mechanism in a data-driven manner to capture the semantic correlations of the interactive body parts. The distance-based graph is constructed by measuring the distance between pairs of body parts to excavate the distance information between interactive body parts. The two interaction graphs are combined to complement each other in a refinement way, making the model suitable for modeling different interactions.

To feed skeleton sequences to the IGFormer, one straightforward solution is 
to transform
each skeleton sequence 
to a pseudo-image 
and divide 
the image 
into a sequence of patches, similar to the manner of  ViT \cite{dosovitskiy2020image}. 
However, this may destroy the spatial relationship among the skeleton joints in each body part, which 
could hinder effective modeling of the interactive body parts for interaction recognition. 
To tackle this problem, we propose a Semantic Partition Module (SPM) to transform the skeleton sequence of each subject 
into a new format, 
i.e., a 
Body-Part-Time (BPT) sequence,
each of which is the representation of one 
body part during a short period. The BPT sequence encodes 
semantic information and temporal dynamics of the body parts, enhancing
the capability of the network for modeling interactive body parts for interaction recognition. 

We summarize the contributions of this paper as follows:
\begin{itemize}
\item We introduce a Transformer-based model named
     IGFormer, which contains a novel GI-MSA module to learn the relationships of the interactive persons 
     from both semantic and distance levels for skeleton-based human interaction recognition.
    \item We introduce a Semantic Partition Module (SPM) transforming each skeleton sequence into a BPT sequence to enhance the modeling of interactive body parts.
    \item We conduct extensive experiments on three challenging datasets 
    and achieve state-of-the-art performance. 
\end{itemize}


\begin{figure*}[h] 
    \vspace{-0.5cm}
    \centering 
    \includegraphics[width=1.0\textwidth]{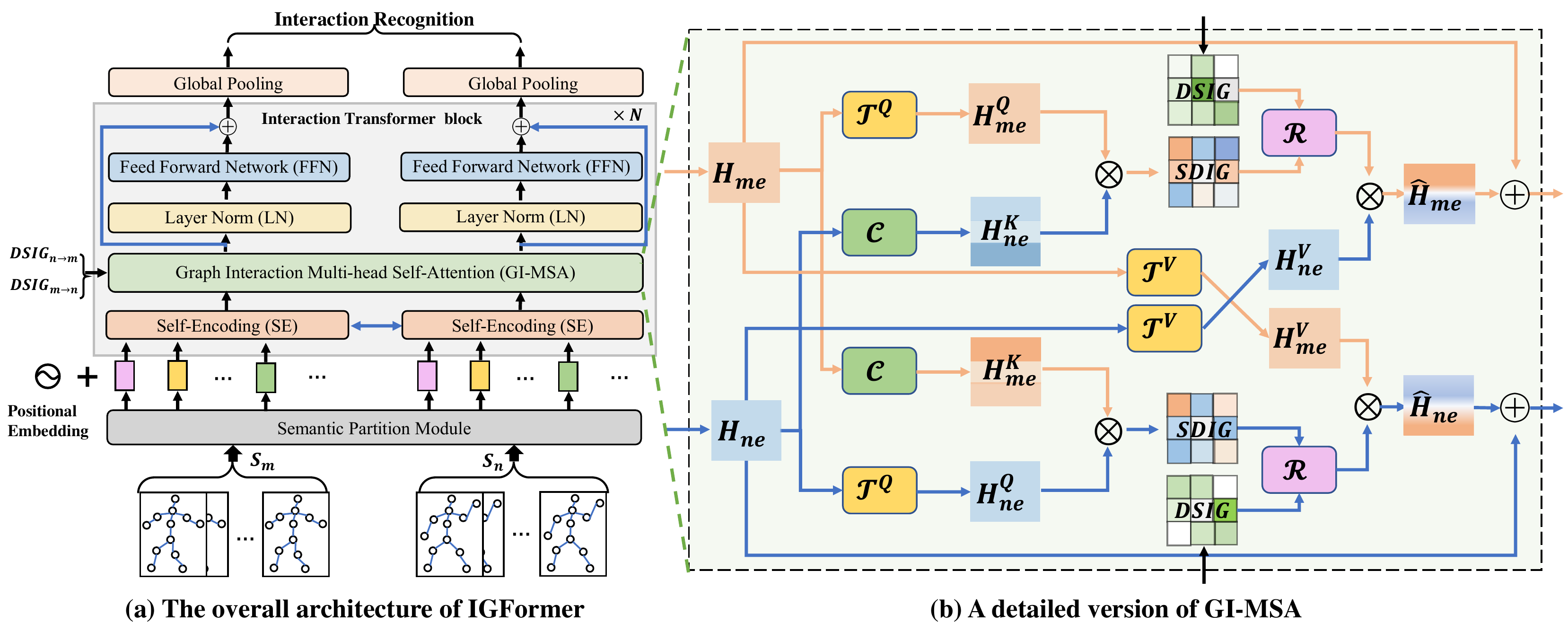} 
    \vspace{-0.4cm}
    \caption{(a) The overall architecture of the proposed IGFormer for skeleton-based human interaction recognition. Given the skeleton sequences of two subjects, they are first fed into the Semantic Partition Module (SPM) to generate two Body-part-time (BPT) sequences. The BPT sequences are then fed into the Interaction Transformer Block (ITB)  for interactive learning. The ITB contains three main components: two self-encoding (SE) modules, the proposed GI-MSA module and two two-layer Feed-Forward Networks (FFN). Finally, a global average pooling followed by a softmax classifier is applied to the outputs of the last ITB to predict interaction labels. (b) The structure of the proposed Graph Interaction Multi-head Self-Attention (GI-MSA) module (DSIG: distance-based sparse interaction graph, SDIG: semantic-based dense interaction graph). } 
    \label{InterFormer}  
    \vspace{-1cm}
\end{figure*}

\section{Related Work}
\subsection{Skeleton-based Action Recognition}
Conventional deep learning-based methods model the human skeleton as a sequence of joint-coordinate vectors \cite{liu2016spatio,Shahroudy_2016_CVPR,du2015hierarchical,song2017end,zhang2017view,li2018independently} or a pseudo-image \cite{liu2017two,ke2017new,kim2017interpretable,li2017skeleton,du2015skeleton}, which is then fed into RNNs or CNNs to predict the actions.
However, representing the skeleton data as a vector sequence or a 2D grid cannot fully express the dependency between correlated joints since the human skeleton is naturally structured as a graph. 
Recently, GCN-based methods \cite{li2019actional,shi2019two,liu2020disentangling} consider the human skeleton as a graph whose vertices are joints and edges are bones and apply graph convolutional networks (GCN) on the human graph to extract correlated features. These methods achieve better performance than RNN- and CNN-based methods, and become the mainstream methods in skeleton-based action recognition. However, these methods consider each person as an independent entity and cannot effectively capture human interaction. In this work, we focus on skeleton-based human interaction recognition and propose to model the interactive relationship of persons from both semantic and distance levels.


\subsection{Human Interaction Recognition} 
Human interaction recognition ~\cite{zhang2012spatio,vahdat2011discriminative,raptis2013poselet} is a sub-field of action recognition. Compared with single-person action recognition, human interaction methods should not only be able to model the behavior of each individual but also capture the interaction between them. Yun et al. \cite{6239234} evaluated several geometric relational body-pose features including joint features, plane features and velocity features for interaction modeling, and found out that joint features outperform others, whereas velocity features are sensitive to noise. Ji et al. \cite{6890714} built poselets by grouping joints that belong to the same body part of each individual to describe the interaction of each body part. Recently, Perez et al.~\cite{perez2021interaction} proposed a two-stream LSTM-based interaction relation network called LSTM-IRN to model the intra relations of body joints from the same person and the inter relations of the joints from different persons. However, LSTM-IRN ignores the distance evolution of body parts, which is considered as an important prior knowledge for human interaction recognition. Different from the above-mentioned methods, we model the interaction relationship of interactive humans as two interaction graphs, which are constructed from the semantic and distance levels respectively to capture the semantic correlation and distance evolution between body parts.

\subsection{Visual Transformer}
Transformer was first proposed in \cite{vaswani2017attention} for machine translation task and since then has been widely adopted in various natural language processing (NLP) tasks. Inspired by the successful application in NLP, Transformer has been applied to the computer vision and demonstrated its scalability and effectiveness in many vision tasks. Vision Transformer (ViT) \cite{dosovitskiy2020image} was the first pure Transformer architecture for image recognition and obtained better performance and generalization than traditional convolutional neural networks (CNNs).
After that, Transformer-based models with carefully designed and complicated architectures have been applied to various downstream vision tasks, such as object detection \cite{zhu2020deformable}, semantic segmentation \cite{zheng2021rethinking} and video classification \cite{arnab2021vivit}. In skeleton-based action recognition, Plizzari et al.~\cite{PLIZZARI2021103219} proposed ST-TR to model the dependencies between joints by substituting the graph convolution operator with the self-attention operator. Different from ST-TR, we focus on human interaction modeling and propose a novel self-attention-based GI-MSA module to model the correlations between body parts of interactive persons.

\section{Interaction Graph Transformer}
One important cue in recognizing human interaction is the interactive body parts. In this section, we introduce an Interaction Graph Transformer (IGFormer), which contains a Graph Interaction Multi-head  Self-Attention (GI-MSA) module to model the interactive body parts at both semantic and distance levels for skeleton-based interaction recognition. The proposed IGFormer is also equipped with a Semantic Partition Module (SPM), which aims at retaining the semantic and temporal information of each body part within the input skeleton sequences for better learning of the interactive body parts. 

The overall architecture of the proposed IGFormer is shown in Fig.~\ref{InterFormer} (a). Given the skeleton sequences of two interactive subjects $\textbf{S}_{m}, \textbf{S}_{n} \in \mathbb{R}^{T \times J \times C}$, where $T$ and $J$ represent the numbers of frames and joints in each frame, respectively, and $C=3$ represents the dimension of the 3D coordinates of each joint, we first feed the two skeletons into the proposed SPM to generate two Body-Part-Time (BPT) sequences, $\textbf{H}_{m}, \textbf{H}_{n}$, which 
are then fed into a stack of \textit{Interaction Transformer Blocks (ITBs)} for interaction modeling. Finally, a global average pooling followed by a softmax classifier is applied to the output of the last ITB to predict the interaction class.  

More specifically, each ITB contains three components including two shared-weight self-encoding (SE) modules, the Graph Interaction Multi-head Self-Attention (GI-MSA) module, and two Feed-Forward Networks (FFN). Each SE module is a standard one-layer Transformer \cite{dosovitskiy2020image}, which aims at modeling the interaction among the body parts within each individual skeleton. The two outputs of the SE are fed into the GI-MSA to model the interactive body parts and generate an enhanced representation for each interactive person. Finally, each output of the GI-MSA is fed to a Layer Normalization (LN) followed by a FFN. We add an addition operation between the output of GI-MSA and FFNs to improve the representation capability of the model. The ITB can be formulated as follows:
\begin{equation}
    \begin{split}
        &\textbf{H}_{me}, \textbf{H}_{ne} = \text{SE}(\textbf{H}_{m}), \text{SE}(\textbf{H}_{n}),\\
        &\hat{\textbf{H}}_{me}, \hat{\textbf{H}}_{ne} =\text{GI-MSA}(\textbf{H}_{me}, \textbf{H}_{ne}),\\
        &\hat{\textbf{H}}_{mo} = \text{FFN}(\text{LN}(\hat{\textbf{H}}_{me})) + \hat{\textbf{H}}_{me}, \\
        &\hat{\textbf{H}}_{no} =
        \text{FFN}(\text{LN}(\hat{\textbf{H}}_{ne})) + \hat{\textbf{H}}_{ne},
    \end{split}
\end{equation}
where $\textbf{H}_{me}$ and $\textbf{H}_{ne}$  denote the outputs of the SE, $\hat{\textbf{H}}_{me}$ and $\hat{\textbf{H}}_{ne}$ denote the outputs of the GI-MSA module, and $\hat{\textbf{H}}_{mo}$ and $\hat{\textbf{H}}_{no}$ are the outputs of the ITB. 

The two SE modules in the first ITB take the Body-Part-Time (BPT) representations of two interactive subjects, i.e, $\textbf{H}_{m}$ and $\textbf{H}_{n}$, as input. The inputs of the SE in the following ITB are the outputs of the previous ITB. In the following subsections, we introduce the proposed SPM and GI-MSA in detail.


\subsection{Semantic Partition Module}
\label{sec:SPM}
\begin{wrapfigure}{r}{0.5\linewidth}
    \vspace{-2cm}
    \centering 
    \includegraphics[width=0.45\textwidth]{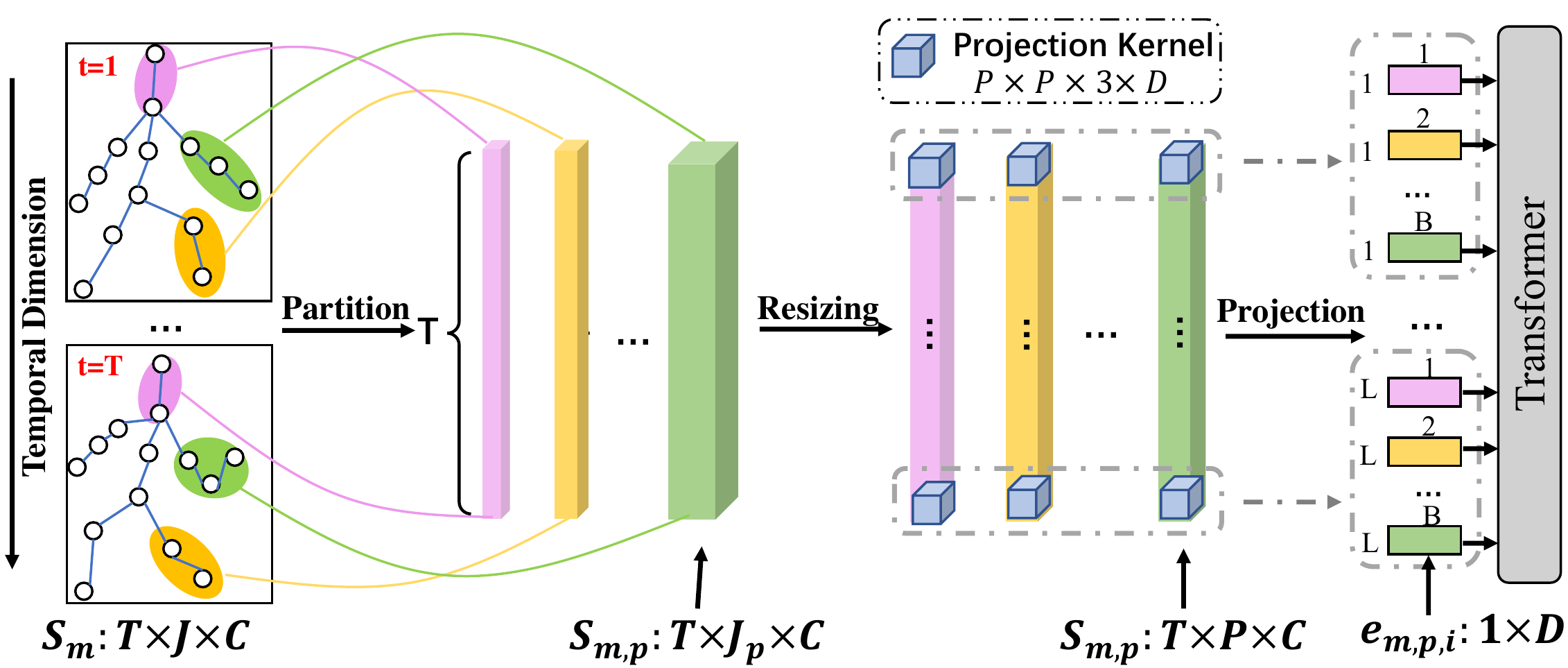} 
    \caption{The proposed Semantic Partition Module (SPM) performs three main operations (i.e., Partitioning, Resizing, and Projection) on the input skeleton sequence to generate its Body-Part-Time (BPT) sequence.}
    \label{SPM}  
    \vspace{-0.8cm}
\end{wrapfigure}

Different from natural 2D images that can be directly divided into a sequence of patches to feed to the Transformer \cite{dosovitskiy2020image}, human skeleton sequences are represented as a set of 3D joints. 
Transforming the 3D skeleton sequences to 2D pseudo-images and passing them through a vision Transformer such as ViT \cite{dosovitskiy2020image} may result in losing the temporal dependency between frames as well as the correlation between joints. To better retain both spatial and temporal information of the skeleton sequences, we propose SPM to transform the skeleton sequence of each subject
into a sequence of BPT. Each element in the BPT is the representation of one body part during a short temporal period. 
The overall architecture of the proposed SPM is shown in Fig.~\ref{SPM}. There are three main steps in the SPM, i.e., partitioning, resizing, and projection, which are explained below.

\par \textbf{Partitioning.} Given
the skeleton sequences of the interactive persons $\textbf{S}_m, \textbf{S}_n\in \mathbb{R}^{T \times J \times C}$, 
we first divide each skeleton sequence into $B$=5 body parts, i.e., \textit{left arm, right arm, left leg, right leg and torso}, according to the natural structure of the human body. 
After the partitioning operation, 
each body part of each subject is represented as $\textbf{S}_{m,p}, \textbf{S}_{n,p} \in \mathbb{R}^{T \times J_{p} \times C}$ , where $p \in B$ and $J_{p}$ is the number of joints of body part $p$. 
\par \textbf{Resizing.} 
Different body parts may have different numbers of joints.
In order to adapt these body parts to the input of the Transformer, we 
adopt the linear interpolation to
resize the spatial dimension $J_{p}$ of all body parts to the same 
dimension $P$,  i.e., 
$\textbf{S}_{m,p}, \textbf{S}_{n,p} \in \mathbb{R}^{T \times J_{p} \times C} \rightarrow \textbf{S}_{m,p}, \textbf{S}_{n,p} \in \mathbb{R}^{T \times P \times C}$, where $p \in B$. After the resizing operation, all $B$ body parts have the same dimension.

\par \textbf{Projection.} The projection operation aims to transform the resized body parts of each person into a BPT sequence to feed to the Transformer. Specifically, we apply a 2D convolution with kernel size of $P \times P$ on $\textbf{S}_{m,p}$ and $\textbf{S}_{n,p}$ to generate 2D feature maps, respectively. The size of each output feature map is $L \times  D$, where $L = \ceil*{(T + 2\times padding - P+1) / stride}$ and $D$ denotes the number of output channels. ``$padding$'' and ``$stride$'' denote the padding size and the stride of the convolutional filter. Each 2D feature map can then be split into a sequence of $L$ steps, where each step is a feature vector of dimension $D$. The projection can be formulated as follows:
\begin{equation}
 \begin{split}
        &
    \textbf{e}_{m,p,1}, \textbf{e}_{m,p,2},...,\textbf{e}_{m,p,L}= \text{Split}(\text{Conv}(\textbf{S}_{m,p})),\\
     & \textbf{e}_{n,p,1}, \textbf{e}_{n,p,2},...,\textbf{e}_{n,p,L}= \text{Split}(\text{Conv}(\textbf{S}_{n,p})),
    \label{Eq_projection}
    \end{split}
\end{equation}
where $\textbf{e}_{m,p,j},\textbf{e}_{n,p,j} \in \mathbb{R}^{D}$ denote the embedding 
of the body part $p$ at 
temporal step $j$ for interactive person $m$ and $n$, respectively. $j \in [1,\cdots, L]$, and $D$ is the dimension of the embedding. $L$ is the number of time steps of each body part. After projection, we concatenate the embedding of all the $B$ body parts step by step for all the $L$ time steps to generate a sequence with  $M=B * L$ time steps. The sequence is referred to as the BPT sequence. As shown in Fig.~\ref{SPM}, the BPT sequence can be considered as a combination of $L$ sub-sequences, each  of which 
is formed by the features of the $B$ body part. We denote the BPT sequences generated from  the  skeleton sequences of the two interactive persons  as $\textbf{H}_m, \textbf{H}_n \in \mathbb{R}^{M \times D}$.
A learnable positional encoding \cite{dosovitskiy2020image} is added to 
$\textbf{H}_m$ and $\textbf{H}_n$ to form the inputs of two shared-weight Self Encoding (SE) modules, which are  standard one-layer Transformers~\cite{dosovitskiy2020image}. The output sequences of SE are denoted as $\textbf{H}_{me},\textbf{H}_{ne} \in \mathbb{R}^{M \times D} $, which
are then fed to the  Graph Interaction Multi-head Self-Attention (GI-MSA) module to model the interactive body parts and generate an enhanced representation for each interactive subject.

\subsection{Graph Interaction Multi-head Self-Attention}
\label{sec:GI-MSA}
To accurately recognize human interaction, one critical cue is the interactive body parts. Considering the semantic 
correspondence and the distance characteristics that may exist in the interactive body parts, we propose a Graph Interaction Multi-head Self-attention (GI-MSA) module to model the interactive body parts as two interaction graphs as shown in Fig.~\ref{InterFormer} (b).
Specifically, GI-MSA contains a Semantic-based Dense Interaction Graph (SDIG) and a Distance-based Sparse Interaction Graph (DSIG).  
The SDIG is learned by exploring the semantic correlations of the interactive body parts in a data-driven manner while
the DSIG is constructed 
based on the prior knowledge that 
the physically close body parts of the interactive persons are generally interactive body parts and should be connected.
With the SDIG and DSIG, the proposed GI-MSA models the interaction relationships of humans from both semantic and distance spaces to capture critical interactive information. 
Finally, the representation of each individual is enhanced by aggregating interactive features from 
the other
person.  

\subsubsection{Semantic-based Dense Interaction Graph}
In order to capture the semantic correlations between interactive body parts of people (e.g., the hands holding the camera of one person and the hand with “yeah” of the other person in the action of ``taking a photo''), we construct a Semantic-based Dense Interaction Graph (SDIG) for each interactive person. 
We take the learning of SDIG of person $m$  (which is denoted as $\text{SDIG}_{m \rightarrow n}$) as an example. As shown in Fig.~\ref{InterFormer} (b),  given the representations of two interactive persons $\textbf{H}_{me},\textbf{H}_{ne} \in \mathbb{R}^{M \times D}$, which are the outputs of the SE module, we first transform $\textbf{H}_{me}$ into the latent space by a linear transformation function $\mathcal{T}^Q$:
\begin{equation}
    \textbf{H}_{me}^Q=\mathcal{T}^Q(\textbf{H}_{me})=\textbf{H}_{me}\textbf{W}^Q,
\end{equation}
where $\textbf{H}_{me}^Q \in \mathbb{R}^{M \times D}$ is the transformed query feature and $\textbf{W}^Q \in \mathbb{R}^{D \times D}$ is the weight matrix. Then, we propose a context transformation function $\mathcal{C}$ to transform the representation of the other person $\textbf{H}_{ne}$ into a high-level space as the key features,
\begin{equation}
    \textbf{H}_{ne}^K=\mathcal{C}(\textbf{H}_{ne})=(\textbf{H}_{ne}+\textbf{H}_{ne}^{tc}+\textbf{H}_{ne}^{sc})\textbf{W}^K,
\end{equation}
where $\textbf{H}_{ne}^K \in \mathbb{R}^{M \times D}$ is the key features. $\textbf{W}^K \in \mathbb{R}^{D \times D} $ is 
the learned weight matrix. 
$\textbf{H}_{ne}^{tc} \in $ and $\textbf{H}_{ne}^{sc}$ are temporal and spatial contexts 
of $\textbf{H}_{ne}$. To compute $\textbf{H}_{ne}^{tc}$ and $\textbf{H}_{ne}^{sc}$, we first compute $\textbf{H}_{ne,p}^{tc}$ and $\textbf{H}_{ne,t}^{sc}$ as follows, which denote the temporal context of each body part $p$ and spatial context at time step $t$ in $\textbf{H}_{ne}$. 
\begin{equation}
    \begin{split}
        \textbf{H}_{ne,p}^{tc} &= \frac{1}{L}\sum_{j=1}^{L}{\textbf{H}_{ne,p,j}},\\ 
        \textbf{H}_{ne,t}^{sc} &= \frac{1}{B}\sum_{i=1}^{B}{\textbf{H}_{ne,i,t}},\\
    \end{split}
    \label{Eq:SDIG_context}
\end{equation}
 where $L$ denotes the temporal steps of each body part in $\textbf{H}_{ne}$, and $B$ is the number of body parts. $\textbf{H}_{ne,p,j} \in \mathbb{R}^D$ and $\textbf{H}_{ne,i,t} \in \mathbb{R}^D$ denote the feature encoding of body part $p$ at time step $j$ and the feature encoding of body part $i$ at time step $t$ in the sequence $\textbf{H}_{ne}$, respectively.
By stacking the temporal context  of all $B$ body parts and repeating $L$ times, 
and repeating the spatial context of each time step $B$ times and stacking the the repetition of  all $L$ time steps,
respectively, we obtain $\textbf{H}_{ne}^{tc}$, $\textbf{H}_{ne}^{sc}  \in \mathbb{R}^{M \times D}$. Finally, $\text{SDIG}_{m \rightarrow n}$ can be obtained by performing the matrix multiplication operation between $\textbf{H}_{me}^Q$ and $\textbf{H}_{ne}^K$:
\begin{equation}
    \text{SDIG}_{m \rightarrow n} = \frac{\textbf{H}_{me}^Q (\textbf{H}_{ne}^K)^\top}{\sqrt{D}},
\end{equation}
where $\top$ is the transpose operation, $\text{SDIG}_{m \rightarrow n} \in \mathbb{R}^{M \times M}$. 
$\text{SDIG}_{n \rightarrow m}$ can be obtained in a similar way, and the learnable weight matrices $\textbf{W}^Q$ and $\textbf{W}^K$ are shared for learning both $\text{SDIG}_{m \rightarrow n}$ and $\text{SDIG}_{n \rightarrow m}$.

\subsubsection{Distance-based Sparse Interaction Graph}
In addition to modeling the interaction relationship from the semantic level, we also compute the distance correlation between body parts of the interactive persons.
The DSIG is a predefined graph and could be constructed in the data pre-processing stage.
The idea of DSIG is to leverage the distance between body parts to construct an adjacency matrix that contains the connection information between body parts of the interactive persons. More specifically, if the distance between two body parts of the interactive persons is small, then the two body parts are connected.
Given the original skeleton sequences of two interactive humans $\textbf{S}_{m}, \textbf{S}_{n} \in \mathbb{R}^{T \times J \times C}$, we first divide the skeleton sequences into $B$ body parts $\textbf{S}_{m,p}, \textbf{S}_{n,p} \in \mathbb{R}^{T \times J_{p} \times C}$ via the same Partitioning process in SPM. To estimate the distance between body parts, we first compute the representations of body parts by averaging the coordinates of joints within each body part:
\begin{equation}
\begin{split}
    \overline{\textbf{S}}_{m,p} &= \frac{1}{J_{p}}\sum_{i=1}^{J_{p}}\textbf{S}_{m,p}[i], 
    i\in J_p,p\in B, \\
    \overline{\textbf{S}}_{n,p} &= \frac{1}{J_{p}}\sum_{j=1}^{J_{p}}\textbf{S}_{n,p}[j], 
    j\in J_p,p \in B,\\
\end{split}
\end{equation}
where $\overline{\textbf{S}}_{m,p}, \overline{\textbf{S}}_{n,p} \in \mathbb{R}^{T \times C}$ are the representations of body part $p$ of two interactive persons respectively. $\textbf{S}_{m,p}[i]$ and $\textbf{S}_{n,p}[j]$ denote the $i$-th joint in $\textbf{S}_{m,p}$ and  the $j$-th joint in $\textbf{S}_{n,p}$, respectively. $J_{p}$ is the number of joints within body part $p$. We downsample the temporal dimension of $\overline{\textbf{S}}_{m,p}$ and $\overline{\textbf{S}}_{n,p}$ from $T$ to $L$, i.e., $\overline{\textbf{S}}_{m,p},\overline{\textbf{S}}_{n,p} \in \mathbb{R}^{T \times C} \rightarrow \overline{\textbf{S}}_{m,p},\overline{\textbf{S}}_{n,p} \in \mathbb{R}^{L \times C}$. 
Then combining the representations of $B$ body parts, we get the representations of two persons $\overline{\textbf{S}}_m,\overline{\textbf{S}}_n \in \mathbb{R}^{M \times C}$ in the distance space, where $M = L \times B$.
$\overline{\textbf{S}}_m,\overline{\textbf{S}}_n$ can be treated as  sequences with $M$ time steps with dimension $C$. Each time step corresponds to a body part at a particular time step of the original sequence. 
For human $m$, we compute the Euclidean distance of each time step $a$ 
in $\overline{\textbf{S}}_m$ ($\overline{\textbf{S}}_{m}[a]$) with each time step $b$ 
in $\overline{\textbf{S}}_n$ ($\overline{\textbf{S}}_{n}[b]$):
\begin{equation}
    \text{A}_{m \rightarrow n}[a,b]=\sqrt{\sum_{c=1}^{C}{(\overline{\textbf{S}}_{m}[a] - \overline{\textbf{S}}_{n}[b])^2}},
\end{equation}
where $a,b \in [1,\cdots,M]$, and $\text{A}_{m \rightarrow n} \in \mathbb{R}^{M \times M}$ records the distance between the body parts of two people. We finally connect each 
time step
$a$ in human $m$ to the $k$ nearest 
time step
in human $n$ to build the $\text{DSIG}_{m \rightarrow n} \in \mathbb{R}^{M \times M}$ as below:
\begin{equation}
\text{DSIG}_{m \rightarrow n}[a,b]=
\left\{
             \begin{array}{lr}
             1,\ \text{A}_{m \rightarrow n}[a,b] <= \text{A}^{k}_{m \rightarrow n}[a]&  \\
             0,\ \text{A}_{m \rightarrow n}[a,b] >\ \ \ \text{A}^{k}_{m \rightarrow n}[a]&  
             \end{array}
\right.
\label{Eq_dsig}
\end{equation}
where $\text{A}^{k}_{m \rightarrow n}[a]$ is the $k$-th smallest value in $a$-th row of $\text{A}_{m \rightarrow n}$. The $\text{DSIG}_{n \rightarrow m} \in \mathbb{R}^{M \times M}$ is built in a similar way to encode the distance between each part of the interactive person $n$ to all body parts of person $m$.

\subsubsection{Interaction-based Feature Generation} 
Given the semantic- and distance-based interaction graphs, we aggregate the interactive information of the graphs with the individual features of the interactive persons to generate an enhanced representation for better interaction recognition as shown in Fig.~\ref{InterFormer} (b).
Specifically, we first transform the input individual representation $\textbf{H}_{ne}$, which is the output of the SE module for person $n$, into the value features $\textbf{H}_{ne}^V$:
\begin{equation}
    \textbf{H}_{ne}^V =\mathcal{T}^V(\textbf{H}_{ne}) = \textbf{H}_{ne}\textbf{W}^V,
\end{equation}
where $\textbf{H}_{ne}^V \in \mathbb{R}^{M \times D}$, and $\textbf{W}^V$ is the weight matrix. Then we perform the matrix multiplication operation on $\textbf{H}_{ne}^V$ and the combination of $\text{DSIG}_{m \rightarrow n}$ and $\text{SDIG}_{m \rightarrow n}$, followed by an addition operation with $\textbf{H}_{me}$ to obtain the interactive representation of person $m$:
\begin{equation}
    \hat{\textbf{H}}_{me} = \mathcal{R}(\text{DSIG}_{m \rightarrow n}, \text{SDIG}_{m \rightarrow n}) \textbf{H}_{ne}^V + \textbf{H}_{me}
\end{equation}
where $\hat{\textbf{H}}_{me} \in \mathbb{R}^{M \times D}$, and $\mathcal{R}$ is the combination function:
\begin{equation}
    \mathcal{R}(\text{DSIG}, \text{SDIG}) = \text{Softmax}(\text{DSIG}_{m \rightarrow n} + \alpha \cdot \text{SDIG}_{m \rightarrow n}),
\end{equation}
where $\alpha$ is a trainable scalar to adjust the intensity of each graph enabling the network to be adaptively adjustable between distance evolution and semantic correlation of body parts. Similarly, $\hat{\textbf{H}}_{ne}$ can be obtained in the same way. 

We define the above steps of generating $\hat{\textbf{H}}_{me}$ and $\hat{\textbf{H}}_{ne}$ from $\textbf{H}_{me}$ and $\textbf{H}_{ne}$ as Graph Interaction Self-Attention (GI-SA), which is formulated as:
\begin{equation}
    \hat{\textbf{H}}_{me},\hat{\textbf{H}}_{ne}=\textbf{GI-SA}(\textbf{H}_{me}, \textbf{H}_{ne}).
\end{equation}

Finally, GI-MSA is defined by considering $h$ attention ``heads'', i.e., $h$ self-attention functions are applied to the input in parallel. Each head provides a sequence of size $M \times d$, where $d = D/h$. The outputs of the $h$ self-attention functions are concatenated to form an $M \times D$ sequence to be fed to the a Layer Normalization (LN) followed by a FFN. The GI-MSA can be formulated as:
\begin{equation}
\begin{split}
    \text{GI-MSA}(\textbf{H}_{me}, \textbf{H}_{ne}) =      &\text{Concat}(\hat{\textbf{H}}_{me,1},...,\hat{\textbf{H}}_{me,h})\textbf{W}^m,\\
    &\text{Concat}(\hat{\textbf{H}}_{ne,1},...,\hat{\textbf{H}}_{ne,h})\textbf{W}^n\\
    \hat{\textbf{H}}_{me,i},\hat{\textbf{H}}_{ne,i}&=\textbf{GI-SA}(\textbf{H}_{me,i}, \textbf{H}_{ne,i}),
\end{split}
\end{equation}
where $h$ is the number of heads, $\hat{\textbf{H}}_{me,i},\hat{\textbf{H}}_{ne,i} \in \mathbb{R}^{M \times d}$ are output representations of $i$-th head of GI-SA, and $\textbf{W}^m,\textbf{W}^n$ are the weight matrices. $\textbf{H}_{me,i},\textbf{H}_{ne,i} \in \mathbb{R}^{M \times d}$ are $i$-th head representations of  $\textbf{H}_{me}$ and $\textbf{H}_{ne}$.
\section{Experiments}
The proposed IGFormer is evaluated on three benchmark
datasets, i.e., SBU~\cite{6239234}, NTU-RGB+D~\cite{Shahroudy_2016_CVPR} and NTU-RGB+D120~\cite{liu2019ntu}, 
and is compared with 
state-of-the-art 
RNN-, CNN- and GCN-based human action and interaction recognition methods, including 
Co-LSTM \cite{zhu2016co}, ST-LSTM~\cite{liu2016spatio}, GCA-LSTM~\cite{8099874}, 2s-GCA~\cite{Liu_2018}, FSNET~\cite{liu2019skeletonbased}, 
VA-LSTM \cite{zhang2017view},
LSTM-IRN~\cite{perez2021interaction}, 
ST-GCN~\cite{yan2018spatial}, AS-GCN~\cite{li2019actional} and CTR-GCN~\cite{chen2021channel}. 
Furthermore, to demonstrate the improvement of the proposed IGFormer over the standard Transformer model, we design a Transformer-based baseline named ViT-baseline, which is a ViT-base \cite{dosovitskiy2020image} model taking the pseudo-image representation of the skeleton sequence as input.


\subsection{Datasets}
\par \noindent \textbf{SBU}~\cite{6239234} is a two-person interaction dataset, which contains eight classes of human interactions including \textit{approaching}, \textit{departing}, \textit{pushing}, \textit{kicking}, \textit{punching}, \textit{exchanging objects}, \textit{hugging}, and \textit{shaking hands}. Seven participants (pairing up to 21 different permutations) performed all eight interactions. In total, the dataset contains 282 short videos. Each video contains 3D coordinates of 15 joints per person at each frame. Following~\cite{6239234} , we use the 5-fold cross validation protocol to evaluate our method.

\par \noindent \textbf{NTU-RGB+D}~\cite{Shahroudy_2016_CVPR} is a large-scale action dataset containing 56,578 skeleton sequences from 60 action classes. Each action is captured by 3 cameras at the same height but from different horizontal angles. 
Each human skeleton contains
3D coordinates of 25 body joints. There are two standard evaluation protocols for this dataset including 1) Cross-Subject, where half of the subjects are used for training and the remaining ones are used for testing, and 2) Cross-View, where two cameras are used for training, and the third one is used for testing. This dataset contains 11 human interaction classes including \textit{punch/slap}, \textit{pat on the back}, \textit{giving something}, \textit{walking towards}, \textit{kicking}, \textit{point finger}, \textit{touch pocket}, \textit{walking apart}, \textit{pushing}, \textit{hugging} and \textit{handshaking}. The maximum number of frames in each sample is 256.  
\par \noindent \textbf{NTU-RGB+D120}~\cite{liu2019ntu} extends NTU-RGB+D with an additional 57,367 samples from 60 extra action classes. In total, it contains 113,945 skeleton sequences from 120 action classes. There are two standard evaluation protocols for this dataset including 1) Cross-Subject, where half of the subjects are employed for training and the rest are left for testing, 2) Cross-Setup, where half of the setups are used for training, and the remaining ones are used for testing. 
In addition to the 11 interaction classes in the NTU-RGB+D,
This dataset contains 15 additional human interaction classes including \textit{hit with object}, \textit{wield knife}, \textit{knock over}, \textit{grab stuff}, \textit{shoot with gun}, \textit{step on foot}, \textit{high-five}, \textit{cheers and drink}, \textit{carry object}, \textit{take a photo}, \textit{follow}, \textit{whisper}, \textit{exchange things}, \textit{support somebody} and \textit{rock-paperscissors}, resulting a total of 26 interaction classes. In both NTU-RGB+D120 and NTU-RGB+D datasets, for samples with less than 256 frames, we repeat the sample until it reaches 256 frames. 

\subsection{Implementation Details}
\par \noindent \textbf{Transformer Architecture.} We use a variant of ViT-Base \cite{dosovitskiy2020image} as the backbone of our proposed IGFormer model. The original ViT-base model contains 12 Transformer layers with the hidden size of 768 (D=768).  The dimension of each MLP layer is four times the hidden size. However, due to the small number of samples in the human interaction recognition datasets, a lighter model is more suitable to avoid overfitting. Therefore, we reduce the number of Transformer layers to 3 (N=3) and initialize them with the pre-trained weights of the first three layers of the ViT-base model. We also remove the classification token (CLS) and adopt the average pooling operation to obtain the final representation from each sequence of patches. We set the patch size $P$ in the Resizing step of SPM to 16 and the stride of convolution in the Projection step to 10, which results in BPT sequences with M=125 for each person in all datasets. In each body part, L equals to 25. $k$ in Eq.~(\ref{Eq_dsig}) is set to 15.
\par \noindent \textbf{Training Details.}
The experiments are conducted on NVIDIA P100 GPU. We adopt SGD algorithm with Nesterov momentum of 0.9 as the optimizer. The initial learning rate is set to 0.01 and is divided by 10 at the $30^{th}$ and $40^{th}$ epochs. The training process is terminated at the $60^{th}$ epoch, batch size is 32.

\subsection{Ablation Study}
In this section, we conduct extensive ablation studies on both NTU RGB+D and NTU RGB+D 120 datasets to validate the effectiveness of the proposed SPM (Section~\ref{sec:SPM}) and GI-MSA (Section~\ref{sec:GI-MSA}) modules.

\begin{table}[tb]
\renewcommand\arraystretch{1.}
	\centering
	\begin{center}
	
	\caption{Performance comparison of different types of inputs and different lengths of the sequences on NTU-RGB+D and NTU-RGB+D 120. ``PI'' denotes ``Pseudo-Image''.}
	\vspace{-0.3cm}
    \scriptsize
	\begin{tabular}{cccccc}
		\hline
		\multirow{2}{*}{\textbf{Input}}& \multirow{2}{*}{\textbf{Length}}& \multicolumn{2}{c}{\textbf{NTU 60 (\%)}} & \multicolumn{2}{c}{\textbf{NTU 120(\%)}}\\
		& & \textbf{X-Sub}& \textbf{X-View}&\textbf{X-Sub}& \textbf{X-Set}\\
		\hline
		\hline
		\multirow{3}{*}{Sequence from PI}&80 & 90.8 &94.1 & 83.2 &84.2 \\ 
		&125& {91.8}& {95.2} & {83.7} &{85.0} \\ 
		&200& 89.7 & 93.9 & 81.9 & 83.1\\ 
		\hline
		\multirow{3}{*}{BPT Sequence} &80& 92.8 & 96.0 & 84.8 &86.1 \\ 
		&125& \textbf{93.6}& \textbf{96.5}  & \textbf{85.4} & \textbf{86.5}\\ 
		&200& 91.9 & 95.1  & 83.8 & 83.9\\ 
		\hline
	\end{tabular}
	\label{ablation-sduty-ntu60}
	\end{center}
	\vspace{-0.5cm}
\end{table}

\begin{table}[tb]
\renewcommand\arraystretch{1.2}
	\centering
	\caption{Performance comparison of different interaction learning methods}
	\vspace{-0.5cm}
	\begin{center}
	\scriptsize
	\begin{tabular}{lcccc}
		\hline
		\multirow{2}{*}{\textbf{Methods}}& \multicolumn{2}{c}{\textbf{NTU 60 (\%)}} & \multicolumn{2}{c}{\textbf{NTU 120(\%)}}\\
	    & \textbf{X-Sub}& \textbf{X-View}&\textbf{X-Sub}& \textbf{X-Set}\\
		\hline
		\hline
		Input Fusion & 90.8 &94.3 & 82.9 &83.8 \\ 
		Late Fusion& 91.2& 94.8 & 83.0 &84.1 \\ 
		IGFormer&\textbf{93.6} & \textbf{96.5} & \textbf{85.4} & \textbf{86.5}\\ 
		\hline
	\end{tabular}
	
	\label{asil}
	\end{center}
	\vspace{-3em}
\end{table}

\par \textbf{Impacts of SPM.} 
We compare two different representations of the skeleton sequences as the input of the proposed IGFormer to validate the effectiveness of the proposed SPM. The first one is \textbf{Pseudo-Image} representation, which have been widely
used in CNN-based models \cite{ke2017new,kim2017interpretable,li2017skeleton} by transforming each 3D skeleton sequence to a 2D pseudo-image. We define the numbers of frames $T$ and joints $J$ of a 
skeleton sequence as the width and height of the image and then perform a linear projection on the image as ViT \cite{dosovitskiy2020image}. 
The second representation is the BPT sequence, which is generated by the proposed SPM. Moreover, skeletons are transformed into the different lengths by changing the stride of convolution projection in ViT and SPM to validate the robustness of the proposed SPM under different input configurations. The experimental results are shown in Table \ref{ablation-sduty-ntu60}. We observe that the BPT representation outperforms Pseudo-Image representation at all three configurations, which validates the effectiveness of the proposed SPM. We also evaluate a baseline that models each skeleton joint as a token of the Transformer sequence and fuses features of two persons, but the performance drops by 2.2\% 
compared with our SPM on X-Sub of NTU-RGB+D.


\par \textbf{GI-MSA versus Input/Late fusion.} We design two interaction learning baselines, i.e.,  \textbf{Input Fusion} and \textbf{Late Fusion}, to compare with our proposed GI-MSA module. The \textbf{Input Fusion} baseline merges the BPT sequences of two subjects to form a single sequence and passes it through a standard Transformer to learn the interactions between two subjects. The \textbf{Late Fusion} baseline feeds the BPT sequences of two subjects individually through a Transformer model to extract their representations, which are then fused to model the interaction. As shown in Table~\ref{asil}, we observe that the performance of both input fusion and late fusion methods are worse than our proposed IGFormer on both datasets, demonstrating the efficacy of the proposed GI-MSA module for interactive learning.

\begin{table}[tb]
	\centering
	\caption{Performance comparison of different components of the proposed GI-MSA module. $sc$ and $tc$ represent spatial context and temporal context in Eq.~(\ref{Eq:SDIG_context}).}
	\vspace{-0.5cm}
	\begin{center}
    \scriptsize
	\begin{tabular}{lcccc}
		\hline
		\multirow{2}{*}{\textbf{Methods}} & \multicolumn{2}{c}{\textbf{NTU 60 (\%)}} & \multicolumn{2}{c}{\textbf{NTU 120(\%)}}\\
	    & \textbf{X-Sub}& \textbf{X-View}&\textbf{X-Sub}& \textbf{X-Set}\\
		\hline
		\hline
		baseline & 90.2 &93.3 &82.1&83.6\\ 
		DSIG & 90.4& 92.9& 82.2& 83.5\\
		SDIG w/o sc & 92.4 &95.5 &84.6&85.4\\ 
		SDIG w/o tc & 92.3 & 95.1 & 84.3& 85.0\\ 
		SDIG & 92.8 & 95.7 & 84.8& 85.5\\ 
		SDIG + DSIG& \textbf{93.6} & \textbf{96.5} &\textbf{85.4} &\textbf{86.5}\\ 
		\hline
	\end{tabular}
	\label{abGIMSA}
	\end{center}
	\vspace{-4.5em}
\end{table}

\begin{wraptable}{r}{0.5\linewidth}
\renewcommand\arraystretch{1.2}
    \vspace{-2.5em}
	\centering
	\caption{Performance comparison of number of ITB layers on NTU-RGB+D and NTU-RGB+D 120 datasets.}
	\vspace{-0.3cm}
	\begin{center}
	\small
	\begin{tabular}{lcccc}
		\hline
		\multirow{2}{*}{\textbf{ITB}}& \multicolumn{2}{c}{\textbf{NTU 60 (\%)}} & \multicolumn{2}{c}{\textbf{NTU 120(\%)}}\\
	    & \textbf{X-Sub}& \textbf{X-View}&\textbf{X-Sub}& \textbf{X-Set}\\
		\hline
		\hline
		2& 92.7&  95.3&  84.0& 85.2 \\ 
		\textbf{3}&\textbf{93.6} & \textbf{96.5} & \textbf{85.4} & \textbf{86.5}\\
		4 & 92.6 & 95.1& 84.2& 85.7\\ 
		5 & 91.9& 94.7& 83.5& 84.8\\ 
		\hline
	\end{tabular}
	
	\label{nITB}
	\end{center}
	\vspace{-2.5em}
\end{wraptable}

\par \textbf{Impacts of SDIG and DSIG.} We evaluate the impacts of different components of the proposed GI-MSA, including SDIG, DSIG, the spatial and temporal context for learning SDIG. Here, we employ IGFormer without GI-MSA module as our baseline. Based on the results in Table \ref{abGIMSA}, we draw three conclusions: (1) Both spatial and temporal context in Eq.~(\ref{Eq:SDIG_context}) are important for learning key contextual features, i.e., the performance drops significantly by removing any of them. (2) The GI-MSA containing only SDIG can improve the performance of human interaction recognition, which validates the effectiveness of the proposed SDIG. (3) The DSIG, which serves as the prior knowledge of human interaction, does not perform well individually but provides extra information for interaction learning, leading to improved performance after being combined with SDIG.

\par \textbf{Impacts of Number of ITB layers.} Our IGFormer is built by stacking several Interaction Transformer Blocks (ITBs) to enhance the capability of interaction modeling. Here, we evaluate the influence of different number of ITBs on the performance of IGFormer. As shown in Table \ref{nITB}, stacking 3 layers of ITB achieves the best results on both NTU-RGB+D and NTU-RGB+D 120. Increasing the number of ITBs degrades the accuracy due to over-fitting problem.

\par \textbf{Impacts of the joint noise on human interaction.} The skeletons in NTU-RGB+D are usually noisy, e.g., some joints are missing. We evaluate the performance of our IGFormer on X-Sub of NTU-RGB+D by adding zero-mean noise to the skeleton sequences. IGFormer achieves 93.6\%, 93.1\%, 92.0\%, 90.4\% accuracy when the standard deviation ($\sigma$) is set to 0cm, 1 cm, 2cm, 4cm, respectively, which demonstrates that IGFormer is robust against the input noise.

\subsection{Comparison with State-of-the-arts}
The experimental results on the interaction classes of SBU, NTU-RGB+D and NTU-RGB+D 120 datasets are shown in Table \ref{acc-NTU60}. The proposed IGFormer achieves state-of-the-art performance compared with other skeleton-based human interaction recognition methods. Benefiting from the proposed SPM and GI-MSA modules, IGFormer outperforms the CNN- and RNN-based methods by a large margin. 
IGFormer also outperforms state-of-the-art GCN-based method, CTR-GCN~\cite{chen2021channel}, by $2.0\%$ and $2.2\%$ on X-Sub and X-View of NTU-RGB+D, and $2.2\%$ and $2.1\%$ on X-Sub and X-set of NTU-RGB+D 120. Compared with the baseline Transformer-based method, ViT-baseline, our IGFormer achieves $3.4\%$ and $3.2\%$ gains on X-Sub and X-View of NTU-RGB+D, and $3.9\%$ and $4.0\%$ gains on X-Sub and X-Set of NTU-RGB+D 120.



\begin{table}[tb]
    \renewcommand\arraystretch{1.2}
	\centering
	\begin{center}
	\caption{Performance comparison on SBU, 
	NTU-RGB+D and NTU-RGB+D 120.
	}
	\vspace{-0.3cm}
    \scriptsize
	\begin{tabular}{lccccc}
		\hline
		\multirow{2}*{\textbf{Methods}}& \multirow{2}*{\textbf{SBU(\%)}}& \multicolumn{2}{c}{\textbf{NTU-RGB+D}} &\multicolumn{2}{c}{\textbf{NTU-RGB+D 120}}\\
		&& \textbf{X-Sub (\%)}& \textbf{X-View (\%)} &\textbf{X-Sub (\%)}& \textbf{X-Set (\%)}\\
		\hline
		\hline
		Co-LSTM \cite{zhu2016co} & 90.4 &- &- &- &- \\
		 ST-LSTM \cite{liu2016spatio} & 93.3 & 83.0 &87.3 & 63.0 &66.6 \\ 
		GCA \cite{liu2017global} & - & 85.9 & 89.0 & 70.6 & 73.7\\ 
		2s-GCA \cite{liu2017skeleton} &94.9 & 87.2 & 89.9 & 73.0 & 73.3\\
		VA-LSTM \cite{zhang2017view} &97.2 &- &- &- &-\\
		FSNET \cite{liu2019skeleton} &- & 74.0 & 80.5 & 61.2 & 69.7\\ 
		LSTM-IRN \cite{perez2021interaction} &98.2 & 90.5 & 93.5 & 77.7 & 79.6\\ 
		\hline
		ST-GCN \cite{yan2018spatial} &- & 83.3 & 87.1 & 78.9 & 76.1  \\
		AS-GCN \cite{li2019actional} &-& 89.3 & 93.0 & 82.9 & 83.7 \\
		CTR-GCN \cite{chen2021channel} &-& 91.6 & 94.3 & 83.2 & 84.4 \\
		\hline
		ViT-baseline &93.1 & 89.7 & 92.5 & 81.5 & 82.5 \\
		\textbf{IGFormer} &\textbf{98.4} & \textbf{93.6} & \textbf{96.5} & \textbf{85.4} & \textbf{86.5} \\
		\hline
	\end{tabular}
	\label{acc-NTU60}
	\end{center}
	\vspace{-3em}
\end{table}

\section{Conclusion}
In this work, we presented IGFomer, which consists of 
a GI-MSA module to model the interaction of persons as graphs. The GI-MSA learns an SDIG and DSIG to capture the semantic and distance correlations between body parts of interactive persons.
We also presented a SPM to transform each human skeleton into a BPT sequence for retaining interactive information of body parts.  The proposed IGFormer outperformed state-of-the-art methods on three datasets.

\section{Acknowledgement}
The research is partially  supported by University of Melbourne Early Career Researcher Grant  (No: 2022ECR008). This research is also partially  supported by TAILOR, a project funded by EU Horizon 2020 research and innovation programme under GA No 952215. This  work is also partially supported by National Research Foundation, Singapore under its AI Singapore Programme (AISG Award No: AISG-100E-2020-065), SUTD Startup Research Grant and MOE Tier 1 Grant.

\clearpage
%
%
\bibliographystyle{splncs04}
\bibliography{egbib}
\end{document}